\documentclass[conference,letterpaper]{IEEEtran}

\IEEEoverridecommandlockouts

\usepackage{cite}
\usepackage{amsmath,amssymb,amsfonts}
\usepackage{algorithmic}
\usepackage{graphicx}
\usepackage{textcomp}
\usepackage{xcolor}
\usepackage{float}
\usepackage{url}
\usepackage{balance}
\usepackage{hyperref}
\usepackage{placeins}
\usepackage{booktabs}
\usepackage{pifont}
\usepackage{indentfirst}

\def\BibTeX{{\rm B\kern-.05em{\sc i\kern-.025em b}\kern-.08em
    T\kern-.1667em\lower.7ex\hbox{E}\kern-.125emX}}

\begin{document}

\title{
    ASTRA: ASV Simulator \\ for Transferable Realistic Autonomy\\[0.5em]}

\author{
\IEEEauthorblockN{Tsimafei Lazouski\textsuperscript{1}, Cody Sheltraw\textsuperscript{2}, Alan Papalia\textsuperscript{1}, and Maani Ghaffari\textsuperscript{1,2}}
\IEEEauthorblockA{\thanks{\textsuperscript{1}Department of Naval Architecture and Marine Engineering, University of Michigan, Ann Arbor, MI \texttt{\{tlazousk,apapalia\}@umich.edu}}
\thanks{\textsuperscript{2}Department of Robotics, University of Michigan, Ann Arbor, MI \texttt{\{cshel,maanigj\}@umich.edu}}
}}

\maketitle
\hypersetup{hidelinks}
\begin{abstract}
Autonomy development for marine autonomous surface vehicles (ASVs) remains challenging because field testing is costly and difficult to reproduce. Existing marine simulation tools provide mature capabilities in vessel dynamics, robotics integration, and visual rendering, but these strengths are emphasized differently across platforms. This paper presents ASTRA, an ASV simulation framework built in NVIDIA Isaac Sim that combines marine vehicle dynamics, photorealistic perception, and Robot Operating System (ROS 2) autonomy interfaces within a common environment.
ASTRA incorporates a Fossen-style marine dynamics model together with geometry-based buoyancy and programmable spectral wave disturbances, enabling non-planar hydrostatic and wave-induced motion. The simulator reproduces the sensor configuration of a physical BlueBoat testbed, including GPS, IMU, LiDAR, and RGB-D sensing exposed through ROS 2. This allows localization, perception, planning, and control software to be developed against interfaces consistent with the physical platform.
The BlueBoat's existing ROS 2 localization and perception modules were ported directly into ASTRA, while targeted zero-propulsion experiments were used to verify the implemented wave-body coupling. The simulated roll response was consistent with the expected non-oscillatory hydrostatic behavior, and heave remained highly coherent with the generated wave field, with approximately unity gain at the dominant wave frequency in the tested sea states. ASTRA is intended as an open-source testbed for marine autonomy development, simulation-based benchmarking, and sim-to-real experimentation.
\end{abstract}

\section{Introduction}
Autonomous surface vehicles (ASVs) have progressed toward practical deployment across marine science and engineering~\cite{zheng2025usv}, with applications including environmental monitoring, hydrographic surveying, and offshore structure inspection~\cite{yuan2023marine,schmidt2020hydrographic,cepeda2023inspection}. Yet, developing reliable autonomy for these platforms remains difficult. Field trials are physically and logistically demanding, often exposing the vehicle only to narrow cases of the operational environment. The challenging test domain also limits rapid iteration, refinement of algorithms, and sensor integration. In robotics, this is often solved through the use of high-fidelity simulations.

\begin{figure}
    \centering
    \includegraphics[ width=1.0\linewidth]{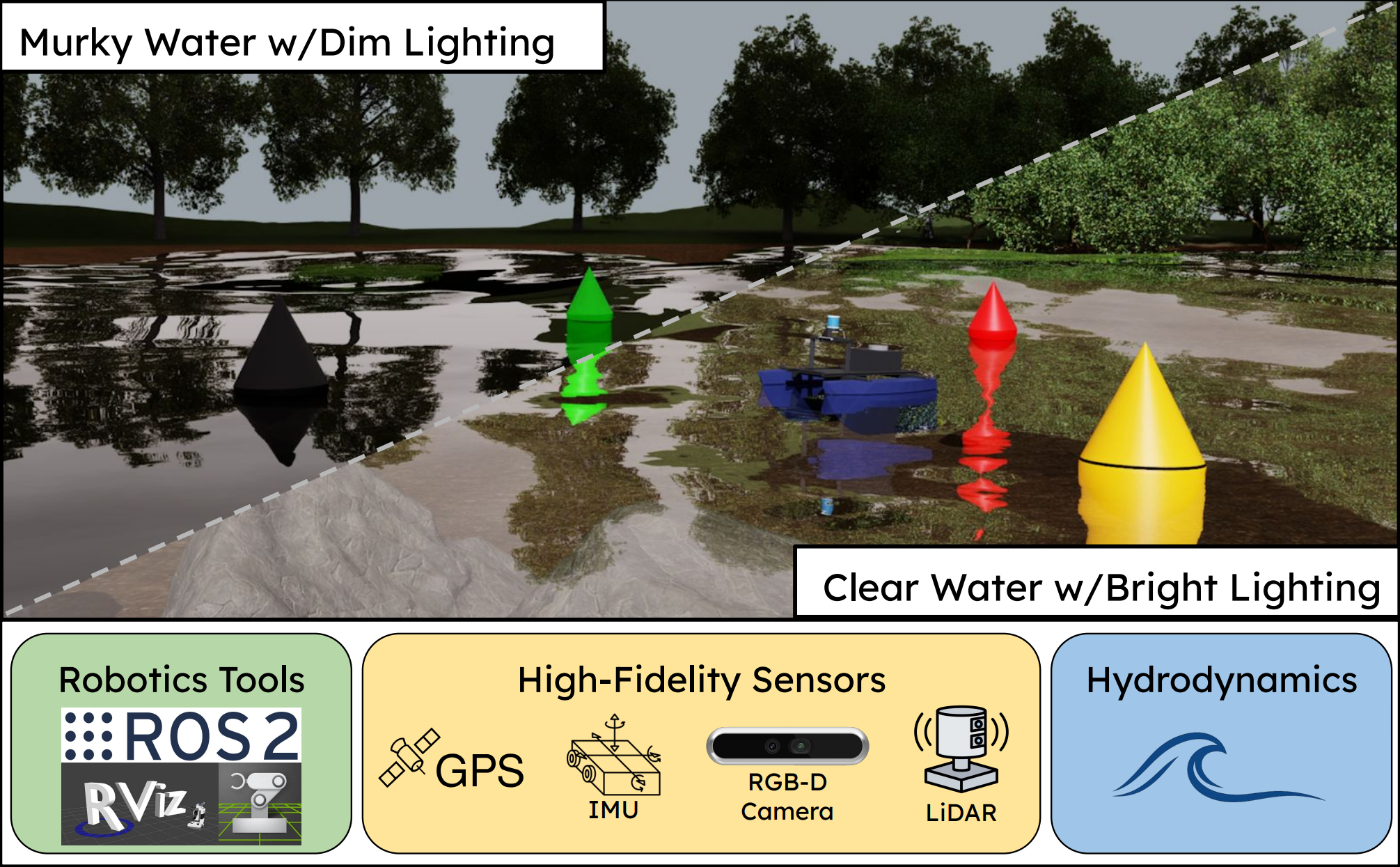}
    \caption{Demonstration of the Isaac Sim simulation environment and the main features supporting sim-to-real transfer research.}
    \label{fig:isaacsim}
\end{figure}

For many years, AirSim~\cite{shah2018airsim} was a strong substrate for aerial vehicles and CARLA~\cite{dosovitskiy2017carla} for autonomous driving. In recent years, however, NVIDIA's Isaac Sim has emerged as a promising core environment for high-fidelity simulation and research. Isaac Sim provides photorealistic rendering, a strong physics engine, sensor models, and established Robot Operating System (ROS 2) development practices. It also naturally opens a path toward reinforcement-learning (RL) applications through Isaac Lab. These strengths have been demonstrated through the successful transfer to hardware of legged-locomotion policies trained in NVIDIA's physics stack~\cite{rudin2022}, and Isaac Lab has consolidated this workflow into a general robot-learning framework~\cite{mittal2023orbit}.  Despite this, the use of Isaac Sim remains limited for marine applications.

While several strong marine simulation solutions are available, there remains a gap in open-source Isaac Sim environments that jointly support surface-vehicle dynamics, wave-body interaction, high-fidelity visual rendering, and ROS 2 autonomy development. Such a simulation environment would allow for joint iteration on the full autonomy stack in a scalable, photorealistic marine surface-vehicle setting. As such, in this work we present \textbf{ASTRA (ASV Simulator for Transferable Realistic Autonomy)}, a simulation framework that couples Isaac Sim's photorealistic rendering and ROS 2 bridge with a Fossen-style 6-degree-of-freedom (6-DOF) dynamics model, a real-time geometry-based buoyancy solver, and an ASV that matches a physical BlueBoat testbed. Packaged as an extension, it is easy to modify for use with a specific platform or test environment. Importantly, Section~\ref{sec:background} positions ASTRA against other marine simulators in more detail. The key contributions of this work are:
\begin{enumerate}
    \item An open-source Isaac Sim framework for full-stack ASV autonomy development, with ROS~2 sensing and a physical BlueBoat counterpart accessible through the \href{https://github.com/UMich-CURLY/ASTRA}{\textbf{ASTRA GitHub repository}}.
    \item A coupled marine simulation model combining Fossen-style 6-DOF dynamics, geometry-based buoyancy, and a programmable spectral wave disturbance, synchronized with the ROS~2 sensor stack.
    \item Field-informed planar dynamics identification and simulator verification, extracted from 29 field-test bags from the physical BlueBoat together with wave-body and hydrostatic verification experiments in Isaac Sim.
\end{enumerate}

\section{Background and Related Work}
\label{sec:background}
Marine robotics simulators span a range of design priorities, including vessel dynamics, perception, middleware integration, and robot-learning workflows. For surface-vehicle autonomy development, we focus on four capabilities that are particularly relevant to ASTRA's design: surface-vehicle support, wave-body interaction, visual realism, and ROS 2 integration. Several mature simulation platforms address these capabilities with different emphases~\cite{marinegym, asvsim, holoocean, stonefish, vrx, rans_asv}. Notably, all of these simulators have some form of ROS 2 support. Table \ref{tab:sim_compare} summarizes these distinctions, while the following discussion positions ASTRA within this existing simulator landscape.
\subsection{Simulation Platform}
There is no single simulator platform that is the best fit for all robotics and marine research. Stonefish ~\cite{stonefish} stands out as a mature, custom-built simulator specifically for marine robotics research. VRX ~\cite{vrx} is built on Gazebo, which provides a strong physics engine and years of hydrodynamic plugin support alongside ROS 2 compatibility. Yet, it lacks the photorealism that Unreal Engine or NVIDIA platforms can provide. Unreal Engine~\cite{holoocean, asvsim} also provides a strong physics engine coupled with high rendering quality. Yet, being a game engine, it lacks a native ROS 2 workflow and the tightly integrated, GPU-parallel robot-learning tooling that NVIDIA's Isaac platforms provide. Isaac Sim targets interactive robotics simulation and provides native sensor and ROS 2 interfaces, while Isaac Lab extends this environment with GPU-parallel tooling for robot-learning applications. This combination of robotics integration and a direct learning workflow motivated ASTRA's use of the NVIDIA simulation stack. 
\subsection{Surface Vehicle Support}
MarineGym~\cite{marinegym} is built exclusively for the underwater domain. Stonefish and HoloOcean were primarily built for underwater robotics research, and while they support surface vehicles, that support remains limited. VRX, ASVSim~\cite{asvsim}, and RANS-ASV~\cite{rans_asv} were explicitly built for surface vehicles. Among the simulators designed specifically for surface vehicles, VRX provides explicit 6-DOF wave-driven vessel motion, whereas ASVSim and RANS-ASV primarily model planar surface-vehicle dynamics. We believe a simulation supporting full 6-DOF is important for studying problems involving wave-induced motion, such as navigation through waves.
\subsection{Waves}
Marine surface autonomy couples perception and control because wave-induced motion perturbs both vehicle dynamics and sensor orientation. Existing simulators address different portions of this problem. VRX provides surface-vessel dynamics driven by a spectral sea state, but its Gazebo-based rendering limits visual realism~\cite{vrx}. HoloOcean provides high-quality Unreal Engine rendering and a broad marine sensing and autonomy framework, but it is built around underwater robotic applications~\cite{holoocean}. Stonefish provides geometry-aware hydrodynamics and FFT-based ocean waves, but the emphasis remains on underwater dynamics and rendering~\cite{stonefish}. ASTRA targets the combination of above-water visual realism, surface-vehicle wave-body coupling, and ROS 2 autonomy interfaces within Isaac Sim. The ``Waves'' column in Table~\ref{tab:sim_compare} denotes explicit wave-field support and does not imply equivalent wave-body fidelity across simulators.
\subsection{Photorealism}

Visual mismatch between simulation and reality is a known contributor to the sim-to-real gap. Prior work has shown that variation in textures, lighting, object appearance, and camera parameters can substantially affect transfer from synthetic to real imagery~\cite{tobin2017}. In marine robotics, recent simulation work similarly emphasizes high-fidelity, physics-based rendering as a means of reducing perception-domain mismatch~\cite{oceansim}. These considerations, together with our prior development experience in VRX, motivated ASTRA's emphasis on perceptual variation in lighting, shoreline texture, and water appearance.

Between Stonefish's custom rendering, Unreal Engine, and NVIDIA's Isaac Sim, we do not identify a single simulation platform that outperforms the others in terms of photorealism, however. We therefore do not treat rendering quality alone as the primary distinction between these platforms. Choosing NVIDIA's Isaac Sim as the development medium was motivated by:
\begin{enumerate}
    \item The possibility of extension to Isaac Lab for RL.
    \item A strong underlying physics engine.
    \item ROS 2 and built-in sensor model support.
    \item Successful usage in the broader robotics community.
\end{enumerate}

\begin{table}[H]
        \centering
    \caption{Positioning of popular marine simulators. \\
    \checkmark: supported, $\sim$: partial / domain-dependent.}
    \label{tab:sim_compare}
    \footnotesize
    \setlength{\tabcolsep}{1.5pt}
    \renewcommand{\arraystretch}{0.8}
    \begin{tabular}{lccccc}
    \toprule
    Simulator & Simulation Platform & Surface & Photorealism & Waves & ROS 2 \\
    \midrule
    VRX~\cite{vrx}        & Gazebo & $\checkmark$ & $\sim$ & $\checkmark$ & $\checkmark$  \\
    Stonefish~\cite{stonefish}  & Custom C++ & $\sim$       & $\sim$   & $\checkmark$ & $\checkmark$     \\
    HoloOcean~\cite{holoocean}  & Unreal Engine & $\sim$     & $\checkmark$ & $\sim$ & $\checkmark$  \\
    MarineGym~\cite{marinegym} & Isaac Sim 4.1.0 & --         & $\checkmark$ & --
    & $\sim$       \\
    RANS-ASV~\cite{rans_asv}   & Isaac Sim 2022.2.1 & $\checkmark$ & $\sim$ & --  & $\checkmark$ \\
    ASVSim~\cite{asvsim}     & Unreal Engine & $\checkmark$ & $\checkmark$ & $\sim$  & $\checkmark$ \\
    \textbf{ASTRA (ours)} & Isaac Sim 5.1.0 & $\checkmark$ & $\checkmark$ & $\checkmark$ & $\checkmark$\\
    \bottomrule
    \end{tabular}
\end{table}

\section{Technical Approach}

The ASTRA extension is built around a small set of cooperating components: the ASV and buoyancy solver, the ocean-surface model, a GNSS sensor node, and a manager for floating environmental debris. All of them advance on the same physics-step callback, so vehicle dynamics, water animation, sensing, and environment interactions stay synchronized within a single step. LiDAR, RGB-D, and IMU sensing are provided directly by Isaac Sim's own ROS 2 sensor prims. The following sections describe each of these pieces.

\subsection{6-DOF Dynamics}
The motions in our simulator are based on Fossen-style surface-vessel dynamics~\cite{fossen2021}. This choice provides a physically meaningful representation of low-speed ASV motion while maintaining a structure compatible with guidance and control methods commonly used for ASVs, where Eq.~\eqref{eq:fossen} provides the governing vehicle model. $M$ is the inertia matrix, $C$ is the Coriolis and centripetal forces matrix, $D$ is the damping matrix, and $g$ represents the hydrostatic term. $\tau$ and $\tau_{wave}$ denote the propulsion and wave-induced forces, respectively.

\begin{equation}
M\dot{\boldsymbol{\nu}} + C(\boldsymbol{\nu})\boldsymbol{\nu}
+ D(\boldsymbol{\nu})\boldsymbol{\nu} + g(\boldsymbol{\eta})
= \boldsymbol{\tau} + \boldsymbol{\tau}_{wave}
\label{eq:fossen}
\end{equation}

The planar surge and yaw damping terms were calibrated from field data as described below, while the remaining damping coefficients were manually selected. For the target low-speed operating regime of the BlueBoat~\cite{blueboat}, two main simplifying assumptions are adopted: $M$ carries rigid-body inertia only, with no separate added-mass matrix, and $D(\nu)$ is diagonal, with no cross-coupling damping terms between degrees of freedom. With that, body-frame velocities and the summed hydrodynamic, thruster, and buoyancy wrenches are applied to the BlueBoat's rigid body every physics step to model the motions. These approximations are intended to strike a practical balance between realism and real-time performance, retaining computational efficiency and scalability.

\subsection{System Identification}
\label{sec:sysid}
The planar damping terms were calibrated using zero-thrust free-decay segments from field data collected with the physical BlueBoat. Surge and yaw were modeled as:

\begin{equation}
X_D = -X_u u - X_{uu}|u|u,
\end{equation}

\begin{equation}
N_D = -N_r r - N_{rr}|r|r,
\end{equation}

A single, constrained non-negative surge coefficient pair was fitted across seven surge decays, and a single yaw pair across seven yaw decays. With the vessel mass of $m = 26.4$~kg and yaw inertia $I_{zz}=4.5$~kg\,m$^2$, the identification yields $X_u=0~\mathrm{N\,s/m}$, $X_{uu}=7.77~\mathrm{N\,s^2/m^2}$, $N_r=2.46~\mathrm{N\,m\,s/rad}$, and $N_{rr}=9.01~\mathrm{N\,m\,s^2/rad^2}$. The fitted models achieved median $R^2=0.94$ in surge and $R^2=0.96$ in yaw. The linear surge coefficient $X_u=0$ converged to its lower, non-negative, bound, indicating predominantly quadratic surge damping over the tested speed range. The remaining sway, heave, roll, and pitch damping terms were not sufficiently excited by these maneuvers and were manually selected.

\subsection{Hydrostatics and water-body interactions}

Rather than prescribing heave, roll, and pitch independently, ASTRA computes buoyancy from a geometry-based approximation of the instantaneous submerged volume. Each BlueBoat pontoon is represented by a lower-resolution proxy mesh that preserves the overall hull shape and waterline profile while reducing the number of vertices queried at each physics step. The resulting submerged geometry determines both the buoyant-force magnitude and its effective point of application. The buoyant force is computed using: 

\begin{equation} 
\mathbf{F}_b = \rho g V_{\mathrm{sub}} \hat{\mathbf{z}},
\label{eq:buoyancy}
\end{equation}

where $\rho$ is the water density and $V_{\mathrm{sub}}$ is the estimated submerged volume. Because $V_{\mathrm{sub}}$ varies with both vessel pose and local water elevation, the same calculation produces hydrostatic restoring behavior in still water and wave-induced motion when the surface changes beneath the vessel. To accomplish that, the following approach is used.

A reference condition is computed offline once from the BlueBoat hull geometry: the at-rest displacement $V_{rest}$ and the per-vertex rest drafts $d_i^{rest}$ it implies.  Then, for a proxy vertex $i$ at world position $(x_i,y_i,z_i)$, the local submerged depth is:

\begin{equation}
d_i = \mathrm{clip}
\left( h_w(x_i,y_i,t)-z_i,\; 0,\; d_i^{max} \right), 
\label{eq:draft}
\end{equation}

where $h_w(x_i,y_i,t)$ is the instantaneous water-surface elevation and $d_i^{max}$ is the maximum allowable submersion associated with that vertex. Relative to a precomputed equilibrium waterline, the total change in draft is:

\begin{equation}
\Delta = \sum_i \left(d_i-d_i^{rest}\right). 
\end{equation}

This deviation is mapped to the submerged volume using a piecewise-linear approximation constrained to $[0,V_{full}]$:

\begin{equation}
V_{sub} = V_{rest} + \Delta 
\begin{cases}
\frac{\displaystyle V_{rest}}{\displaystyle\sum_i d_i^{rest}} , & \Delta \leq 0,
\\
\frac{\displaystyle(V_{full}-V_{rest})} {\displaystyle\sum_i (d_i^{max}-d_i^{rest})}, & \Delta > 0, 
\end{cases}
\label{eq:vsub}
\end{equation}

Here, $V_{rest}$ is the submerged volume at the nominal waterline and $V_{full}$ is the full proxy-hull volume. With that, the buoyancy wrench is applied at the draft-weighted centroid of the submerged vertices. Consequently, asymmetric submersion naturally produces restoring roll and pitch moments, allowing simulation of 6-DOF motions.

Finally, the same water-surface query is used for floating environmental objects. Smaller objects such as branches, buoys, and bottles use a reduced set of sampling points and a simplified draft-based buoyancy calculation at reduced computational cost. This allows the ASV and surrounding debris to respond consistently to a common time-varying water surface without requiring computationally expensive fluid simulation.

\subsection{Wave spectrum and disturbance realization}
\label{sec:waves}

The water surface is synthesized on the GPU using a modified version of NVIDIA's Omniverse ocean-deformation implementation~\cite{nvidia_omni}, whose TMA-based directional spectral formulation follows Horvath~\cite{horvath2015}. ASTRA retains this spectral wave-synthesis approach and couples the resulting time-varying elevation field directly to the geometry-based buoyancy model. This coupling allows the rendered wave field to generate heave, roll, and pitch disturbances on the ASV.

The underlying JONSWAP spectral density is:

\begin{equation}
S_J(\omega) = \gamma^{r(\omega)} \frac{\alpha g^2}{\omega^5} \exp \left[ -\frac{5}{4} \left( \frac{\omega_p}{\omega} \right)^4 \right],
\label{eq:jonswap}
\end{equation}

where $\gamma$ controls peak enhancement and 

\begin{equation}
r(\omega) = \exp \left[ -\frac{(\omega-\omega_p)^2} {2\sigma^2\omega_p^2} \right],
\end{equation}

with

\begin{equation}
\sigma = 
\begin{cases}
0.07, & \omega \leq \omega_p,
\\
0.09, & \omega > \omega_p.
\end{cases}
\end{equation}

The spectral scale parameter $\alpha$ and peak angular frequency $\omega_p$ are updated from the configured wind speed $U$ and fetch length $F$ according to:
\begin{equation}
\alpha = 0.076 \left( \frac{U^2}{gF} \right)^{0.22},
\qquad \omega_p = 22 \left| \frac{g^2}{UF} \right|^{1/3}
\label{eq:jonswap_params}
\end{equation}

To account for finite water depth, the JONSWAP spectrum is modified using the TMA correction:

\begin{equation}
S(\omega) = \Phi(\omega_H)S_J(\omega),
\label{eq:tma}
\end{equation}

where

\begin{equation}
\omega_H = \mathrm{clip} \left( \omega\sqrt{d/g}, 0, 2.0 \right), 
\end{equation} 

\begin{equation}
\Phi(\omega_H) = 
\begin{cases}

\tfrac{1}{2}\omega_H^2, & \omega_H \leq 1, 
\\
1-\tfrac{1}{2}(2-\omega_H)^2, & \omega_H > 1. 
\end{cases}
\label{eq:tma_Phi}
\end{equation}

Here, $\mathrm{}{d}$ is the configured water depth and $\mathrm{\gamma = 3.3}$ by default. To realize the spectrum, it is discretized into $N=1000$ frequency components spanning wavelengths from $0.1$ to $250$ m. Each component is assigned a random phase and an amplitude proportional to $\sqrt{2S(\omega_i)\Delta\omega }$, and the components are then superposed into a periodic traveling-wave profile. The spectral model therefore determines the relative frequency content of the surface, while the overall wave amplitude is controlled independently by a user-defined scale parameter. The current implementation does not yet infer a physical $H_s$ directly from wind speed and fetch. 

Directionality is introduced by distributing the resulting profile across 128 propagation directions, weighted by a configurable spreading function centered on the wind direction. The directional components are then sampled onto the water mesh to obtain the time-varying elevation field:
\begin{equation}
h_w = h_w(x,y,t)
\end{equation}

Although spectral energy is depth-corrected using the TMA formulation, the current implementation uses the deep-water dispersion relation $\omega^2=gk$. Incorporating the finite-depth relation, $\omega^2=gk\tanh(kd)$, is left for future refinement.

This elevation field provides the coupling between the spectral wave model and the vessel dynamics. At every physics step, $h_w(x,y,t)$ is evaluated beneath the hull vertices in Eq.~\eqref{eq:draft}. Changes in local elevation alter the estimated submerged volume and center of buoyancy, thereby generating heave, roll, and pitch responses as waves pass beneath the vessel. The hydrostatic and wave-induced terms in Eq.~\eqref{eq:fossen} are therefore not evaluated as two independent force models. In still water, the geometry-based buoyancy calculation provides the hydrostatic restoring wrench represented by $g(\boldsymbol{\eta})$. When $h_w(x,y,t)$ varies in space and time, the corresponding change in that wrench produces the wave-induced disturbance represented by $\boldsymbol{\tau}_{wave}$. Both effects are obtained from the same submerged-volume calculation. 

\subsection{Autonomy Stack}
\subsubsection{Sensors}
Autonomy stack compatibility is essential to this work. Isaac Sim contains robotics interfaces providing LiDAR, RGB-D, and IMU outputs, making it well-suited for this goal. Since Isaac Sim does not natively supply a GNSS sensor, collected real-world data were used to develop a custom configuration that mimics RTK-GPS capabilities of the Septentrio mosaic-X5 sensor. 

\subsubsection{Simulation Environment}
 The developed simulator includes a diverse inland water scene intended to expose the autonomy stack to variation in lighting, visual environmental features, and shoreline structure. Floating algae patches, branches, buoys, shoreline vegetation, and underwater rocks are added from the Isaac Sim asset library, enriching the scene overall. Water transparency can be toggled on or off, as shown in Figure \ref{fig:isaacsim}, and scene lighting can be adjusted. These features allow for flexible simulation of test environments in varying perceptual conditions.

\section{Demonstration}

To demonstrate the functionality of our simulator, we aim to answer the following questions:
\begin{enumerate}
\item[\textbf{(Q1):}] Do the simulated waves induce meaningful motions on the ASV?
\item[\textbf{(Q2):}]Can existing autonomy modules from the physical testbed be used directly in simulation?
\item[\textbf{(Q3):}]Is ASTRA fast enough for interactive development?
\end{enumerate}

\textit{\textbf{(Q1): Wave and Buoyancy Verification.}}

To verify the wave-body interactions described earlier, we ran zero-propulsion tests directly in ASTRA at varying wave and ASV starting conditions, checking that the wave-buoyancy coupling chain (surface elevation $\rightarrow$ submerged volume $\rightarrow$ buoyancy force $\rightarrow$ vessel motion) behaves as intended and that the buoyancy model's roll dynamics are consistent with a hydrostatics-only prediction from the same hull geometry. These experiments do not claim the dynamics are faithful to the real world and are intended only to demonstrate that the water mesh and buoyancy estimator meaningfully drive the boat's motion. 

\subsubsection{Roll response}
We tested free roll decay in still water as a consistency check. Since no measured roll field data exist, we use several test-specific assumptions. A static analysis of the hull mesh gives a roll stiffness of $K_\phi=220.5$~N m/rad. For the test, a roll inertia of $I_{xx}=1.0$~$\mathrm{kg\,m^2}$ and total roll damping $c = K_p + c_{buoy} + c_{PhysX} = 30.0$~N m\,s/rad were used. Here, $K_p$ denotes the linear roll-damping coefficient in $D(\nu)$, $c_{\mathrm{buoy}}$ is the damping contribution introduced by the buoyancy solver, and $c_{\mathrm{PhysX}}$ is the additional rigid-body angular damping applied by the physics engine. Because all three act simultaneously in simulation, all three are included in the effective damping used for this consistency check. This gives $\zeta \approx 1.01$, corresponding to a slightly overdamped, non-oscillatory response. Two trials released from $\pm8^\circ$ returned upright without overshoot, consistent with this qualitative prediction. Because the roll inertia and non-planar damping terms have not been calibrated from field data, we treat this result as an internal consistency check rather than a quantitative validation of the physical BlueBoat's roll dynamics. 
\subsubsection{Spectrum realization}
The roll test verifies the hydrostatic restoring behavior of the buoyancy model, but it does not verify that the surface realizes the prescribed JONSWAP/TMA spectrum of Section~\ref{sec:waves}. To show that, surface elevation was sampled for 600 s at a fixed point, independent of the vessel, at three sea states spanning the tested period range, and compared against the commanded spectrum using Welch's method~\cite{welch1967}, which estimates the power spectral density by averaging the periodogram over $K$ overlapping windowed segments of length $N$, using window function $w[n]$ and normalization $U=\frac{1}{N}\sum_n w[n]^2$:

\begin{equation}
\hat{S}(\omega) = \frac{1}{K}\sum_{k=0}^{K-1} \frac{1}{N U}\left| \sum_{n=0}^{N-1} w[n]\,x_k[n]\,e^{-j\omega n \Delta t} \right|^2,
\label{eq:welch}
\end{equation}

\begin{figure}
    \centering    
    \includegraphics[width=0.85\linewidth]{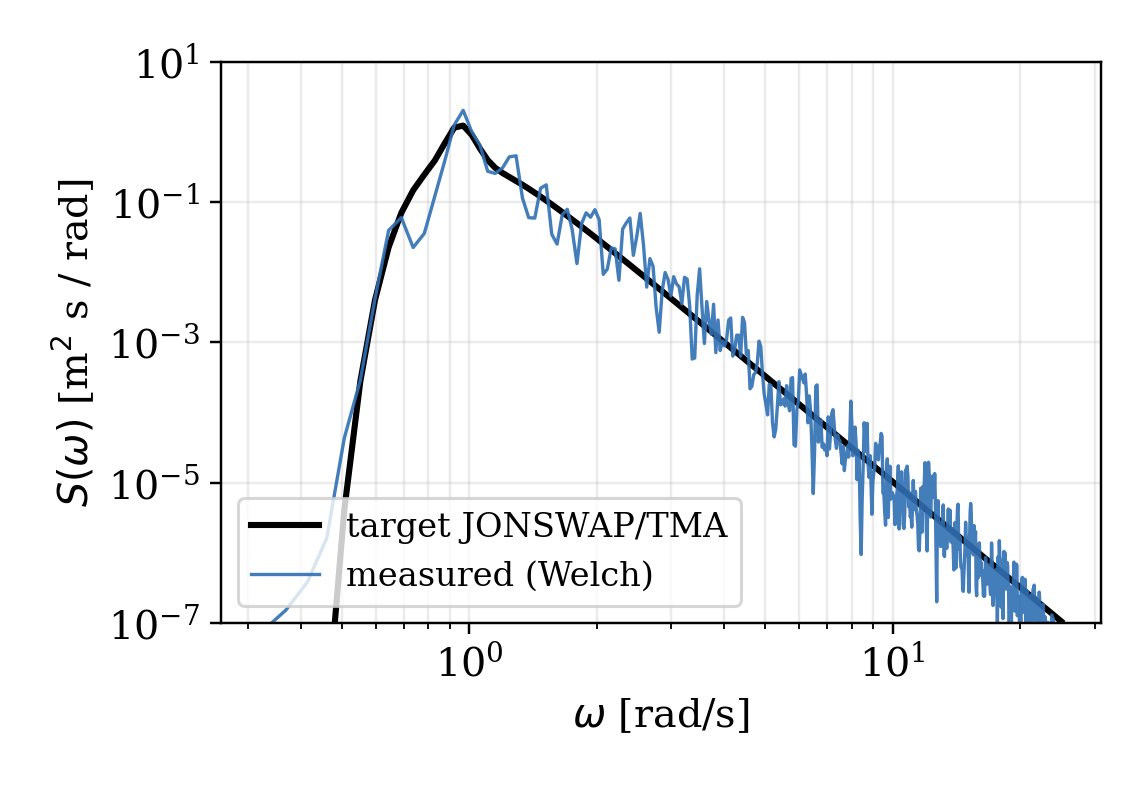}
    \caption{Realized vs. target JONSWAP/TMA spectrum, strong sea state.}
    \label{fig:spectrum_validation}
\end{figure}

With directional spreading disabled, the realized spectra reproduce the target peak and overall spectral shape across all three tested sea states. Figure~\ref{fig:spectrum_validation} shows the strong condition as a representative case. 

\subsubsection{Wave-body response}
To verify vessel motion is actually driven by the local wave field, we computed the frequency-domain coupling between wave elevation and heave for the mild condition of Table~\ref{tab:severity} ($T_p=2.63$~s, $H_s=0.29$~m, zero propulsion). 


At the measured dominant wave frequency, the two signals show a coherence of $0.996$ and a heave RAO of $1.02$~m/m, and a phase lag of $14^\circ$. The same coupling analysis at the strong condition ($H_s=0.61$~m) gives a coherence of $0.992$ and an RAO of $1.00$~m/m, and a phase lag of $20^\circ$. A still-water control run over the same duration showed no measurable heave. Across conditions, heave, roll, and pitch RMS responses grow monotonically with significant wave height, computed from the realized surface-elevation time series (Table~\ref{tab:severity}), with no evidence of saturation over the tested range.

\begin{table}[h]
    \centering
    \caption{RMS vessel response vs. sea state, zero propulsion, $T_p=2.63$~s.}
    \label{tab:severity}
    \footnotesize
    \setlength{\tabcolsep}{4pt}
    \renewcommand{\arraystretch}{0.7}
    \begin{tabular}{lrrr}
    \toprule
    Condition & $H_s$ [m] & Heave [m] & Roll / Pitch [deg] \\
    \midrule
    Still  & 0.00 & 0.0   & 0.00 / 0.00 \\
    Mild   & 0.29 & 0.074  & 3.66 / 3.70 \\
    Strong & 0.61 & 0.149 & 6.72 / 6.77 \\
    \bottomrule
    \end{tabular}
\end{table}

\textit{\textbf{(Q2): Autonomy-Stack Portability}}

\begin{figure}[h]
    \centering
    \includegraphics[width=0.48\textwidth]{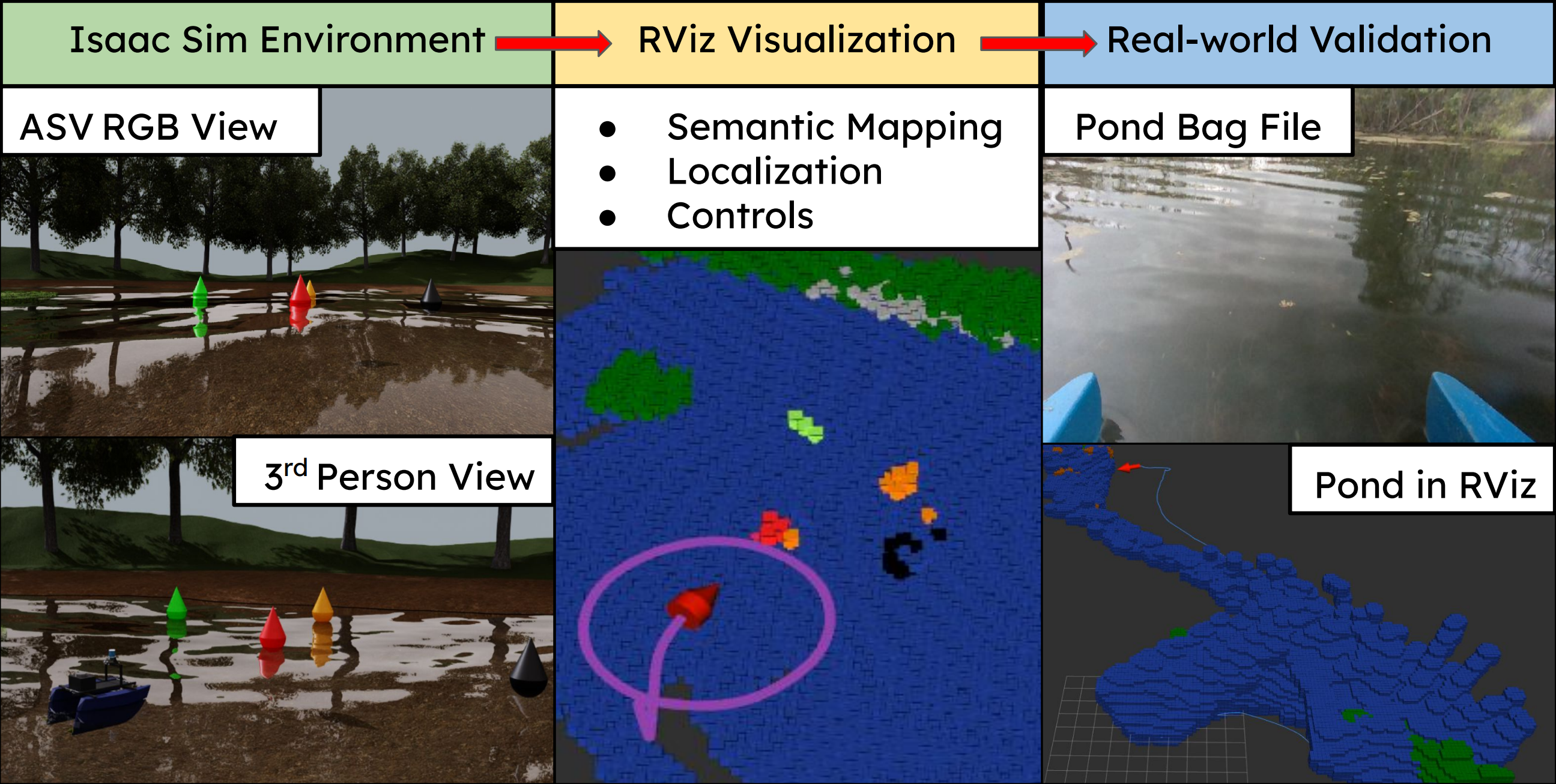}
    \caption{On the left are the 3rd-person and RGB views of the Isaac Sim stage and the semantic map in RViz built from its localization, LiDAR, and RGB-D inputs. On the right is the same ConvBKI pipeline, with the same weights and configuration, run offline on real-world data from the physical BlueBoat.}
    \label{fig:sim2real}
\end{figure}

To evaluate ASTRA's sensor and autonomy interfaces, we ported existing BlueBoat modules without changing their ROS 2 interfaces. GPS-DRIFT~\cite{gpsdrift} is an Invariant Extended Kalman Filter (InEKF) localization algorithm that fuses IMU and GPS data and provides pose, twist, and path ROS 2 messages. ConvBKI~\cite{convbki}, a probabilistic semantic mapping module, tests the usability of the LiDAR and camera while fusing the perception algorithms with localization. Both modules are directly ported from the physical BlueBoat, where they have already demonstrated general functionality. The Lie-algebraic Model Predictive Controller~\cite{Junwoo2023} used on the real vehicle is architecturally compatible with these same interfaces. Figure~\ref{fig:sim2real} summarizes the simulated autonomy stack pipeline and its evaluation on recorded real-world data.

\subsection{Localization}
GPS-DRIFT~\cite{gpsdrift} fuses proprioceptive IMU propagation with GPS position corrections in a symmetry-preserving InEKF and runs directly against ASTRA's GNSS and IMU interfaces, publishing the same pose, twist, and path messages used on the physical BlueBoat at 61.8 Hz. Against Isaac Sim ground truth over a 189 m trajectory, it achieves a position RMSE of less than 3 cm.

\subsection{Semantic Mapping}
We use ConvBKI~\cite{convbki}, a convolutional Bayesian kernel inference network for real-time probabilistic semantic mapping, as a stress test for the combined use of LiDAR, RGB-D, and pose (Figure~\ref{fig:sim2real}, left and center panels). It consumes point clouds, images, and pose at the sensor rate and accumulates them into a persistent map over a run. Map updates were published at 2.2 Hz, producing a spatially coherent map of the stage. This demonstrates that ASTRA's LiDAR and RGB-D outputs can drive ConvBKI while the mapping pipeline consumes pose estimates from GPS-DRIFT.

To probe transfer, this same ConvBKI pipeline, with the same weights and configuration, was then run offline on a recorded ROS 2 bag of real-world sensor data collected by the physical BlueBoat on the pond that ASTRA models. The resulting real-world semantic map is qualitatively consistent with the scene structure recovered in the simulated run.

\textit{\textbf{(Q3): Simulator Performance.}}

Table~\ref{tab:perf} collects the observed rates. The pond stage maintains stable playback at approximately 35 FPS on a single NVIDIA RTX 4090 GPU with the full sensor suite (LiDAR, RGB-D camera, IMU, GNSS), buoyancy solver, and the described autonomy stack active, which is enough to run the ROS 2 modules at rates useful for interactive development.

\begin{table}[H]
    \centering
    \caption{Observed runtime rates with demonstration stack active.}
    \label{tab:perf}
    \footnotesize
    \setlength{\tabcolsep}{4pt}
    \renewcommand{\arraystretch}{1.0}
    \begin{tabular}{llr}
    \toprule
    Quantity & Source & Rate \\
    \midrule
    Stage playback      & Isaac Sim, full sensor suite & 35 FPS \\
    Pose estimate       & GPS-DRIFT & 61.8 Hz \\
    Semantic map update & ConvBKI & 2.2 Hz \\
    \bottomrule
    \end{tabular}
\end{table}

\subsection{Limitations}
\label{sec:limitations}
ASTRA currently provides field-informed planar (surge, yaw) damping but has not been calibrated against physical heave, roll, and pitch response. As such, the roll-decay check above is a consistency result against a placeholder inertia, not a real-vehicle validation. Semantic-mapping transfer is similarly qualitative, since no labeled ground truth exists yet for the real pond. Finally, the wave model here is verified for repeatable, physically consistent disturbance generation rather than field-validated seakeeping prediction or a calibrated sea state.

\section{Discussion and Conclusions}
\label{sec:conclusion}
ASTRA provides an Isaac Sim-based ASV environment combining ROS 2 sensing, field-informed marine dynamics, geometry-based buoyancy, and programmable wave disturbances. Field data improved the simulator's planar damping model, while hydrostatic and wave-response experiments verified the intended vessel-water coupling. The existing BlueBoat autonomy modules, running directly against ASTRA's ROS 2 interfaces, demonstrated the usability of the platform at interactive rates. Future work will focus on calibrating the remaining 6-DOF dynamics and propulsion model, along with extension to  Isaac Lab for reinforcement learning research applications.

\section*{Acknowledgment}
This material is based upon work supported by the Air Force Office of Scientific Research under award number FA9550-25-C-B010 in the amount of \$144,500.

\FloatBarrier 
{\balance
\bibliographystyle{ieeetr}
\bibliography{bibliography.bib}
}

\end{document}